\documentclass[11pt]{article}
\usepackage[final]{acl}   
\usepackage[T1]{fontenc}
\usepackage[utf8]{inputenc}
\usepackage{times}
\usepackage{tikz}
\usetikzlibrary{positioning,fit,arrows.meta,calc}
\usepackage{booktabs}
\usepackage{array}
\usepackage{amssymb}
\usepackage{url}
\usepackage{textcomp}
\usepackage{graphicx}

\newcommand{\scheme}{LLMersion}
\newcommand{\sys}{LLMersion-1}
\newif\ifmeasured
\measuredfalse

\newcommand{\ck}{$\square$}

\begin{document}

\title{LLMersion: A Local-First AI Agent Framework for\\
Low-Cost Home Language Learning toward Educational Equity}

\author{%
\setlength{\tabcolsep}{9pt}%
\renewcommand{\arraystretch}{1.12}%
\begin{tabular}{ccc}
\begin{tabular}[t]{@{}c@{}}
\textbf{Qiming Guo} \\
{\normalfont\small Dept.\ of Computing Sciences} \\
{\normalfont\small Texas A\&M University--Corpus Christi} \\
{\normalfont\small Corpus Christi, TX, USA} \\
{\normalfont\small\texttt{qguo2@islander.tamucc.edu}}
\end{tabular} &
\begin{tabular}[t]{@{}c@{}}
\textbf{Jinwen Tang} \\
{\normalfont\small Dept. of Electrical Engineering and Computer Science} \\
{\normalfont\small University of Missouri} \\
{\normalfont\small Columbia, MO, USA} \\
{\normalfont\small\texttt{jt4cc@umsystem.edu}}
\end{tabular} &
\begin{tabular}[t]{@{}c@{}}
\textbf{Xingran Huang} \\
{\normalfont\small Dept.\ of Computer Science and Engineering} \\
{\normalfont\small University of California, Riverside} \\
{\normalfont\small Riverside, CA, USA} \\
{\normalfont\small\texttt{xhuan230@ucr.edu}}
\end{tabular} \\ \noalign{\vskip 20pt}
\begin{tabular}[t]{@{}c@{}}
\textbf{Hung-Yu Lin} \\
{\normalfont\small Dept.\ of Computing Sciences} \\
{\normalfont\small Texas A\&M University--Corpus Christi} \\
{\normalfont\small Corpus Christi, TX, USA} \\
{\normalfont\small\texttt{hlin5@islander.tamucc.edu}}
\end{tabular} &
\begin{tabular}[t]{@{}c@{}}
\textbf{Yafu Zhong} \\
{\normalfont\small Dept.\ of Mechanical and Aerospace Engineering} \\
{\normalfont\small University of South Florida} \\
{\normalfont\small Tampa, FL, USA} \\
{\normalfont\small\texttt{yafu@usf.edu}}
\end{tabular} &
\begin{tabular}[t]{@{}c@{}}
\textbf{Xiatian Zhuang} \\
{\normalfont\small Dept.\ of Civil and Environmental Engineering} \\
{\normalfont\small University of Missouri} \\
{\normalfont\small Columbia, MO, USA} \\
{\normalfont\small\texttt{xzrhb@umsystem.edu}}
\end{tabular}
\end{tabular}}

\maketitle

\begin{abstract}
Artificial intelligence helps education most where an essential provision
has been rationed by cost. For language learners that provision is a
teacher's voice, which binds listening, reading, speaking, and writing
into one act. Published evidence shows why most learners lack it, from a global shortage of 44 million teachers to heavy household
tutoring bills, and why technology has not substituted for it:
computer-assisted language learning proved effective but narrow,
applications presuppose connectivity 2.6 billion people lack, and One
Laptop per Child's randomized evaluation found that hardware without
capable software teaches nothing. We distill eight difficulties and four
binding constraints, and argue that small open-weight models dissolve the
last: a complete four-skill stack now fits a \$200-class laptop and, on
community measurements, generates at the pace speech is consumed, for
about one US cent of electricity per study hour. We therefore propose
LLMersion, a scheme for AI for education that runs entirely at home, over
the learner's own documents, with an AI-written, AI-understood,
AI-updated codebase anyone can customize; present \sys{}, a released
open-source prototype (\url{https://github.com/QM378/LLMersion}); and
outline the vision of a private learning agent.
\end{abstract}


\section{Introduction}
\label{sec:intro}

Where artificial intelligence helps education most is not where instruction
is already abundant, but where an essential provision has been rationed by
cost. One-to-one guidance is the canonical example: it is the most effective
form of instruction we can measure and, in Bloom's words, ``too costly for
most societies to bear on a large scale'' \citep{bloom1984}. This paper is
about one sharply defined piece of that rationed provision; about the
technologies that have tried, for half a century, to substitute for it; and
about a scheme, not merely a system, for supplying it at the cost of
electricity.

The piece in question is familiar to anyone who learned English primarily
through text. Many learners read technical English fluently yet cannot follow
the same material by ear, and freeze when asked to say its vocabulary aloud.
The cause is structural: solitary reading severs the sounds of a language
from its written form. A learner alone with a book practices exactly one
binding, eyes to meaning, while the bindings that speech requires,
sound to text and articulation to sound, are never exercised. In natural
acquisition a person supplies the missing link: a teacher reads the text
aloud, so the learner hears the words while seeing them; models
pronunciation, so new vocabulary arrives with its spoken form; listens to the
learner speak, closing the loop from perception to production; and corrects
what the learner writes. Being read to is itself a measurable engine of
vocabulary growth \citep{elley1989}, and comprehensible input drives
acquisition generally \citep{krashen1985}. Learners with access to such a
voice acquire listening, reading, speaking, and writing as one bound skill;
learners without it are left with text alone, and their skill profile shows
it.

That much has long been argued qualitatively. The first half of this paper
establishes its factual scale from published evidence. The human provision is
scarce at its source: the world needs 44 million additional teachers by 2030
\citep{unesco2024globalteachers}, and in several major school systems the
English teachers who do exist test below the proficiency they are required
to teach: more than half of $\sim$27{,}000 tested Indonesian teachers
scored at the lowest TOEIC band \citep{coleman2009rsbi}. Where a teacher is
present, oral attention is diluted across classes averaging 21 students in
OECD primary schools \citep{oecd2025eag} and larger elsewhere. The consequence
is stratification: fluent English carries a 34\% hourly wage premium for men
in India \citep{azam2013english}, and when China added a listening component
to its college-entrance English examination, the rural-urban gap in college
access widened by roughly 30\%; the authors' stated mechanism is that
listening and speaking skills require exactly the extracurricular resources
disadvantaged students lack \citep{li2024english}. Households answer by
buying the provision privately: private supplementary education reached 29.2
trillion won in Korea in 2024 with English the leading subject
\citep{kosis2025private}; tutoring households in China spent a mean of RMB
8{,}438 per year in a nationally representative 2019 survey
\citep{wei2024ciefr}; private coaching absorbs 11.8\% of Indian household
education spending \citep{mospi2019nss75}; and 77\% of Vietnamese
upper-secondary students attend private tutoring \citep{dang2007tutoring}.

Prior technological approaches, examined next, reveal an instructive contradiction. Computer-assisted language learning is not ineffective;
the strongest meta-analysis finds it at least as effective as instruction
without technology \citep{grgurovic2013meta}. But it has been narrow:
hand-built tutoring systems shipped fixed lessons; automatic speech
recognition, before deep learning, roughly doubled its error rate on
non-native speech, making oral feedback unreliable for precisely the
learners who needed it \citep{mccrocklin2026asr}; commercial applications
presuppose subscriptions and connectivity that 2.6 billion people do not
have \citep{itu2024facts}. The contradiction reaches its clearest form in the
One Laptop per Child program: a randomized evaluation across 318 Peruvian
schools raised computers per student tenfold and moved no mathematics or
language score \citep{cristia2017olpc}. Hardware was never the missing
ingredient; capable software running on it was.
Table~\ref{tab:problems} synthesizes the published evidence into eight difficulties, the
mechanism that sustains each, the prior technological answer, and its limit.

What has changed, and what the second half of this paper builds on, is
that the missing software has become small enough to own. Open-weight
language models of 1--3 billion parameters occupy 0.8--2\,GB quantized
\citep{hf2024llama32gguf}, a complete synthesis--recognition--translation
stack adds under 2\,GB more \citep{hexgrad2025kokoro}, and community
measurements report generation at the pace speech and text are consumed
\citep{brysbaert2019reading} on hardware as cheap as a \$200-class laptop or an \$80 single-board
computer \citep{raspberrypi2023pi5}. At the 2024 average US residential
electricity price \citep{eia2025residential}, an hour of study on such a
device costs on the order of one US cent. The provision that Bloom priced
beyond societies' reach can now, in this one domain, be run at home.

We therefore propose \scheme{}, a \emph{scheme} rather than only a
system: (P1) a web
interface backed entirely by local, free, lowest-hardware-cost models; (P2)
the complete listening-reading-speaking-writing loop, played in the
teacher's order (read aloud to the learner, discuss what was read, listen
to the learner speak, correct what the learner writes); (P3) the learner's own
documents and self-built vocabulary as first-class input, not a fixed
curriculum; and (P4) a codebase that is AI-written, AI-understood, and
AI-updated, so that customization does not require the professional
engineering that priced earlier tools out of reach. We present \sys{}, a
released open-source reference instance implementing all four principles,
together with a designed but not yet administered feedback instrument for
future community evaluation.

Our contributions:
(1)~an evidence-based analysis of learner needs, associated costs,
and prior technological approaches, distilled into a reusable
problem taxonomy
(Table~\ref{tab:problems}, Sections~\ref{sec:need}--\ref{sec:constraints});
(2)~an analysis identifying the four constraints that have bound low-cost,
personalized, equitable language education, and the evidence that the last
of them has newly dissolved (Sections~\ref{sec:constraints}--\ref{sec:scheme});
(3)~a constraint-driven, local-first educational agent architecture: four
design principles, each answering one identified constraint, including
the AI-maintained-code principle (Section~\ref{sec:scheme}); and
(4)~a released, continuously maintained reference instance
(code: \url{https://github.com/QM378/LLMersion}; companion narration
tool: \url{https://github.com/QM378/llmersion-narrator},
Section~\ref{sec:narrator}) with a designed
feedback instrument and pre-stated expectations for future evaluation
(Section~\ref{sec:instance}); the prototype remains under active development,
and as feedback data accumulates we will continue to upgrade it and will
report findings of value.

Section~\ref{sec:need} examines learner needs; Section~\ref{sec:tech} examines prior technological approaches and their limitations; Section~\ref{sec:constraints}
distills the binding constraints; Section~\ref{sec:scheme} states the scheme and
its feasibility; Section~\ref{sec:instance} describes the reference instance;
Section~\ref{sec:outlook} discusses limitations and the outlook toward a private
learning agent; Section~\ref{sec:conclusion} concludes.

\begin{table*}[t]
\caption{Eight difficulties of second-language acquisition under resource
scarcity, the mechanism sustaining each, technology's prior answer, its
documented limit, and this scheme's response. Sections
\ref{sec:need}--\ref{sec:constraints} develop each row with sources.}
\label{tab:problems}
\centering
\footnotesize
\setlength{\tabcolsep}{3pt}
\renewcommand{\arraystretch}{1.15}
\begin{tabular}{@{}p{2.7cm} p{3.4cm} p{2.4cm} p{3.3cm} p{3.3cm}@{}}
\toprule
\textbf{Difficulty} & \textbf{Why it persists} & \textbf{Prior answer} &
\textbf{Its limit} & \textbf{This scheme} \\
\midrule
Absolute teacher scarcity &
Demand outstrips trained supply; attrition nearly doubled 2015--2022
\citep{unesco2024globalteachers} &
Broadcast courses, MOOCs, apps &
Presuppose connectivity 2.6B people lack \citep{itu2024facts} &
The teacher's four roles as software on hardware already owned \\
Teachers' own oral proficiency &
Teachers formed under grammar/exam paradigms; majorities test below required
levels \citep{coleman2009rsbi} &
Broadcast native audio &
One-way; no feedback on the learner &
Synthesized model voice over any chosen text \\
Individual speaking time rationed &
Classes of 21+ students; teacher talk dominates \citep{oecd2025eag} &
ASR pronunciation drills &
Pre-deep-learning ASR $\sim$doubled errors on L2 speech
\citep{mccrocklin2026asr} &
Unlimited private spoken turns with feedback \\
Method locked to written exams &
High-stakes tests reward reading and grammar \citep{liu2024tblt} &
Test-preparation software &
Reinforces the lock &
All four skills exercised over the same document \\
Reading outruns speech &
Without oral input/output, print decouples from phonology
\citep{costello2021reading} &
Audiobooks, subtitles &
Fixed library; no interaction with the learner's material &
Read-aloud, then spoken discussion, of the learner's own text \\
Proficiency tracks wealth and geography &
Oral practice is bought privately; +34\% wage premium in India
\citep{azam2013english}; listening tests widened China's rural-urban gap
\citep{li2024english} &
Subscription applications &
Recurring fees are a regressive filter &
One-time commodity hardware; near-zero marginal cost \\
Families pay heavily for the gap &
Shadow education: 29.2T won in Korea \citep{kosis2025private}; RMB 8{,}438/yr
per tutoring household in China \citep{wei2024ciefr} &
Cheaper or freemium apps &
$\sim$8\% of monthly users convert to paid \citep{duolingo2024tenk}; content
remains fixed &
Free, open weights; the learner's own materials \\
Cheap hardware without capable software &
Distributing devices alone moved no test score (randomized evidence)
\citep{cristia2017olpc} &
Device-distribution programs &
Hardware was never the bottleneck &
Small open models supply the missing software locally \\
\bottomrule
\end{tabular}
\end{table*}

\section{The Need: Second-Language Acquisition under Resource Scarcity}
\label{sec:need}

This section establishes, from published evidence, why the four-skill binding
is out of reach for most learners. We proceed from what the human provision
does (Section~\ref{sec:whatvoice}), to the scarcity of that provision at its source
(Section~\ref{sec:scarcity}), to the stratification it produces
(Section~\ref{sec:strat}), to the skill profile it leaves behind
(Section~\ref{sec:mute}), to what families pay to compensate (Section~\ref{sec:pay}), and
finally to why unassisted text study cannot substitute
(Section~\ref{sec:whytext}). Every quantitative claim below is traced to a primary
report or peer-reviewed study; where only vendor or community figures exist,
we say so explicitly and attribute them.

\subsection{What the Human Provision Does}
\label{sec:whatvoice}

One-to-one instruction remains the most effective form of teaching we can
measure: Bloom found two-sigma gains under mastery tutoring
\citep{bloom1984}, and a meta-analysis of modern randomized evaluations
confirms tutoring among the most reliably effective educational
interventions \citep{nickow2020}. For language in particular, the tutor's
contribution is concrete and enumerable. First, the voice: a teacher reads
text aloud while the learner sees it, and being read to is itself a
measurable engine of vocabulary growth \citep{elley1989}, consistent with the
broader finding that comprehensible input drives acquisition
\citep{krashen1985}. Reading while listening, the self-administered form of
the same act, improves comprehension, reading speed, and vocabulary
retention across a sustained line of classroom studies
\citep{brown2008rwl,chang2009,chang2011,changmillett2014,changmillett2015},
and a meta-analysis finds that glossing plus aural support reliably aids
incidental vocabulary learning \citep{yanagisawa2020}. Second, the model:
new vocabulary arrives with its spoken form, so the learner's phonological
representation is built at first contact rather than repaired later
\citep{nation2001}. Third, the ear: the teacher listens to the learner speak
and corrects production, the step shadowing research identifies as binding
perception to articulation \citep{kadota2019}. Fourth, the pen: the teacher
corrects what the learner writes. A learner with access to this provision
exercises all four bindings on the same material; a learner without it
exercises one.

\subsection{Scarcity at the Source}
\label{sec:scarcity}

The provision is scarce before any question of quality arises. UNESCO's 2024
global report finds the world needs 44 million additional teachers to reach
universal primary and secondary education by 2030, 15 million of them in
sub-Saharan Africa; closing the gap is costed at roughly \$120 billion per
year; and global primary-teacher attrition nearly doubled between 2015 and
2022, from 4.62\% to 9.06\% \citep{unesco2024globalteachers}. In the OECD's
2024 teaching survey, the share of principals reporting that shortages of
qualified teachers hinder instruction rose again over 2018 levels
\citep{oecd2025talis}.

For English the shortage is compounded by a second deficit: many of the
teachers who do exist cannot themselves model the language they must teach.
In Indonesia, when roughly 27{,}000 teachers in the country's
international-standard schools took the TOEIC, more than half scored in the
lowest band of the 990-point scale and under one percent reached
professional working proficiency \citep{coleman2009rsbi}; a decade later,
reviewers arguing for a B2 minimum standard concluded the picture had not
materially changed \citep{renandya2018english}. Comparable
below-benchmark proportions have been reported for state English teachers
elsewhere in Southeast Asia \citep{renandya2018english}.

Even a proficient teacher cannot give what a large class dissolves. Public
primary classes average 21 students across the OECD
\citep{oecd2025eag}, and are substantially larger in many of the systems
where English carries the highest stakes. The arithmetic is unforgiving and
requires no study to state: in a class of fifty, a full hour divided
equally into individual speaking turns yields barely a minute per learner,
before any teaching happens in between. The one activity a tutor uniquely
provides, listening to this learner speak and responding, is precisely the
activity a classroom is least able to ration out.

\subsection{Stratification: Proficiency Tracks Wealth and Geography}
\label{sec:strat}

Because the oral provision must be bought where schools cannot supply it,
measured ability follows money. In India, fluency in English carries a 34\%
hourly wage premium for men relative to no English, and even partial English
a 13\% premium; the return to fluency is comparable to that of completing
secondary school \citep{azam2013english}. The market prices exactly the
skill the poor cannot practice.

China offers an unusually clean identification of the mechanism. A study
in the \textit{Journal of Development Economics} analyzing administrative
records for 16 million college applicants exploits the staggered introduction of a listening component,
worth 20\% of the English subject score, into the national college-entrance
examination: rural candidates' English percentile ranks fell by about two
points, the rural-urban gap in college access widened by roughly 30\%, and
the authors attribute the effect to listening and speaking requiring
extracurricular resources that disadvantaged students lack
\citep{li2024english}. A test of the bound skill, added to an exam of the
written one, converted a resource gap directly into an admissions gap.

Cross-national proficiency indices tell a compatible story but must be
handled with care: the most-cited ranking, the EF English Proficiency
Index, is computed from millions of self-selected online test takers and is
explicitly not population-representative \citep{efepi2025method}; we
therefore use it for no quantitative claim in this paper.

\subsection{The Read-but-Cannot-Speak Profile}
\label{sec:mute}

Where oral practice is scarce and examinations are written, pedagogy adapts
rationally, and the adaptation leaves a signature. Studies of Chinese
university English teaching document the persistence of grammar-translation
and examination-oriented practice and the difficulty of implementing
communicative and task-based approaches against class sizes and test
pressure \citep{liu2024tblt}. The equilibrium is self-reinforcing:
high-stakes written tests reward the skills that large classes can drill,
so reading and grammar advance while listening and speaking wait; learners
emerge able to parse text they could never follow by ear. The pattern is
common enough across East Asian systems to have earned vernacular names,
and its cause is structural rather than individual: it is the expected
output of instruction in which the sound of the language is rarely heard
and almost never produced.

\subsection{What Families Pay}
\label{sec:pay}

Households do not accept the gap passively; they buy the missing provision
on a private market whose scale is documented by official statistics
(Table~\ref{tab:spending}). South Korea's national survey puts private
supplementary education at 29.2 trillion won in 2024, with 80\% of students
participating and English the single largest subject at 248{,}000 won per
month per participating student in 2023 \citep{kosis2025private}. In China,
a nationally representative 2019 survey found tutoring households spending
a mean of RMB 8{,}438 per year (median 3{,}000), with participation at
31.4\% in urban areas against 14.1\% in rural ones and 46.2\% in first-tier
cities \citep{wei2024ciefr}; the state judged the sector consequential
enough to shut most of it down by decree in 2021
\citep{china2021doublereduction}, closing over 90\% of private tutoring
entities \citep{lyu2025doublereduction}, a policy episode that confirms the
scale of the demand rather than eliminating it. In India, private coaching
absorbs 11.8\% of household education expenditure, and the national
participation rate rose to 27\% by 2025 \citep{mospi2019nss75,mospi2025cmse}.
In Vietnam, 31\% of primary, 56\% of lower-secondary, and 77\% of
upper-secondary students attend private tutoring \citep{dang2007tutoring}.
In Japan, the ministry of economy's industry survey of English-conversation
schools, a top-firm sample covering most of the market, annualizes to
roughly \textyen 61 billion for 2024 \citep{meti2024eikaiwa}. The pattern is
global: shadow education has been documented as a major household
expenditure across Asia, Africa, and Europe
\citep{bray2009,bray2012adb,bray2021africa,bray2021europe}. Whatever else
these figures show, they price the need: families worldwide pay
substantial, recurring, regressive sums for a provision that is, at its
core, a competent voice with time for one learner.

\begin{table*}[t]
\caption{What families pay: one official or peer-reviewed anchor per
country. Figures are for private supplementary education; scope notes
matter and are part of the claim.}
\label{tab:spending}
\centering
\footnotesize
\setlength{\tabcolsep}{3.5pt}
\renewcommand{\arraystretch}{1.15}
\begin{tabular}{@{}p{1.4cm} p{6.4cm} p{6.0cm}@{}}
\toprule
\textbf{Where} & \textbf{Anchor (data year)} & \textbf{Scope note} \\
\midrule
Korea & 29.2T won total; English top subject at 248k won/mo (2023--24)
\citep{kosis2025private} & Official national survey \\
China & RMB 8{,}438/yr mean per tutoring household; 46.2\% participation in
first-tier cities (2019) \citep{wei2024ciefr} & Nationally representative
survey; sector largely closed by 2021 decree
\citep{china2021doublereduction,lyu2025doublereduction} \\
India & 11.8\% of household education spending; participation 27\% by 2025
\citep{mospi2019nss75,mospi2025cmse} & Official NSS rounds; no per-subject
breakdown \\
Vietnam & 77\% of upper-secondary students tutored (survey era)
\citep{dang2007tutoring} & Peer-reviewed household data \\
Japan & $\approx$\textyen 61B/yr, conversation schools (2024)
\citep{meti2024eikaiwa} & Ministry survey; top-firm sample, not full census \\
\bottomrule
\end{tabular}
\end{table*}

\subsection{Why Text-Only Self-Study Cannot Compensate}
\label{sec:whytext}

The obvious zero-cost substitute, studying alone from text, fails for a
reason learning science can state precisely. Language presented in paired
modalities is encoded through complementary verbal and imaginal channels
\citep{paivio1986,mayer2009}; the phonological loop of working memory,
rehearsing the sound of a new word, is a core device of vocabulary learning
\citep{baddeley1998}; and offering multiple, learner-chosen modes of
representation is a recognized principle of accessible instructional design
\citep{cast2018}. None of this rests on the contested claim that learners
should be matched to a fixed sensory ``style,'' for which evidence is
lacking \citep{pashler2008}; the point is that combined presentation serves
everyone, and a book supplies only one mode.

Reading, moreover, does not smuggle the sound in. Electrophysiological
evidence shows that skilled prelingually deaf readers recognize written
words without relying on phonological coding at all
\citep{costello2021reading}: reading competence can develop fully decoupled
from sound. We cite this as a mechanistic demonstration, not as a study of
hearing learners; but the implication for them is direct. A learner who
only ever reads English can build a large orthographic and semantic
vocabulary while the phonological representations that listening and
speaking require are never formed, or are formed privately and wrongly and
then rehearsed. Text-only study does not merely progress slowly toward the
bound skill; it progresses confidently away from it, which is why the
profile of Section~\ref{sec:mute} is stable. The missing ingredient is not
effort or material. It is a voice attached to the text, and an ear
attached to the learner.

\section{Technology's Compensations and Their Contradictions}
\label{sec:tech}

Technology has tried to substitute for the rationed voice for half a
century. The record is not one of failure; it is one of narrowness, and the
distinction matters for what follows.

\subsection{A Half-Century of Computer-Assisted Language Learning}

The standard periodization traces behavioristic CALL (mainframe
drill-and-practice), communicative CALL, and integrative CALL through the
1960s to the 1990s \citep{warschauer1998computers}, a framing later
critiqued as chronologically loose but still the field's reference map
\citep{bax2003call}. Each generation moved closer to the tutor's role;
none reached the voice-and-ear loop of Section~\ref{sec:whatvoice}.

\subsection{Effective, but Narrow}

The honest starting point is a concession: computer-assisted instruction
worked. The strongest meta-analysis of comparative studies finds
technology-supported language instruction at least as effective as
instruction without technology, and superior in the most rigorously
designed studies \citep{grgurovic2013meta}. Any argument that dismisses
five decades of CALL as ineffective is contradicted by the evidence, and
we do not make it.

What the evidence does support is a claim about the shape of that
effectiveness. First, content was fixed. Parser-based intelligent tutoring
systems required a hand-built lexicon, morphological analyzer, parser, and
error grammar per language; the exemplary Japanese tutor ships one fixed
sequence of twenty-four lessons for every user \citep{nagata2009robosensei},
and porting the approach to a new language was a multi-year research
project in itself \citep{amaral2011icall}. No such system could ingest the
learner's own documents. Second, the speaking loop was unreliable exactly
where it was needed. Goodness-of-pronunciation scoring
\citep{wittyoung2000} depends on speech recognition, and recognition
degrades sharply on the accented speech of learners: dictation systems
transcribe native speakers at 90.25\% accuracy but Chinese-L1 speakers at
72.45\%, and their accuracy does not correlate with human judgments of
intelligibility, so the tool cannot tell a learner whether they would be
understood \citep{mccrocklin2026asr}. Even a modern recognizer roughly
triples its word error rate between native and non-native children's read
speech in benchmark comparisons \citep{wills2023nonnative}. A feedback channel
that is least accurate for its intended users cannot carry the tutor's
ear.

\subsection{The Application Era}

Consumer applications scaled access dramatically and inherited both
limits. The flagship efficacy study, commissioned by the vendor, estimated
34 hours of app study as equivalent in placement-test gain to a first
college semester, from a cohort in which 156 participants began and 88
finished \citep{vesselinov2012duolingo}. Independent scholarship read the
result narrowly: the study compared the app to nothing
\citep{krashen2014duolingo}, an independent semester-long case study found
high attrition and gains concentrated among already-committed learners
\citep{loewen2019duolingo}, and the commercial-app efficacy literature as a
whole draws on a strikingly small circle of commissioned authors
\citep{jiang2021duolingo}. None of this makes the applications useless;
it locates them. They deliver fixed curricula well to learners who
persist, and the market leader's own annual report shows how the economics
work: 116.7 million monthly users, of whom 9.5 million, about 8\%, pay
\citep{duolingo2024tenk}. The product that is free at the margin is not the
tutor's loop; the fuller product is a subscription.

Language learning has now entered the LLM era, and the early literature
maps both the promise and the boundary. Reviews of ChatGPT for language
teaching find genuine affordances for explanation, text generation, and
practice alongside accuracy and pedagogy concerns
\citep{kohnke2023chatgpt}, classroom studies frame the model as a
complement to, not a substitute for, the teacher \citep{jeon2023llmedu},
and a systematic review of speech-recognition chatbots argues that large
language models finally make open spoken practice plausible, the very
capability the pre-LLM assistants of the previous generation delivered
only in constrained form \citep{jeon2023beyond,dizon2020assistants}.
What has not changed is the access model: every deployed instantiation of
this literature routes through a metered cloud service, from ChatGPT
itself \citep{openai2023plus} to the market leader's premium tier built
on a frontier model \citep{duolingo2023max}. The pedagogy took the LLM
turn; the economics did not. Constraint C2 survives the LLM era intact,
which is precisely the gap the scheme of Section~\ref{sec:scheme} addresses by
taking the LLM turn local.

\subsection{What Compensation Costs}

Each compensation channel carries a recurring price. Human oral practice
online is sold by the hour: major marketplaces list lessons from \$4 and
typically \$4--40 per hour \citep{italki2026pricing,preply2026pricing}, a
real reduction on in-person tutoring that is still a wage-indexed,
per-hour cost. Frontier conversational AI, the first technology to offer
something like open spoken interaction, is metered or subscribed: the
reference consumer tier costs \$20 per month \citep{openai2023plus}, with
inference served from datacenters. For a learner in a high-income country
these are modest sums; against the income distributions of
Section~\ref{sec:strat} they reproduce exactly the gradient the provision was
supposed to flatten.

\subsection{The Access Contradiction}

All of the above presupposes connectivity, and the presupposition excludes
the population with the most to gain. The ITU's 2024 statistics count 2.6
billion people offline, a third of humanity, 1.8 billion of them rural;
in low-income countries only 27\% of people use the internet, and a fixed
broadband subscription costs on the order of a third of average monthly
income \citep{itu2024facts}. Even the device is a barrier: an entry-level
internet-capable handset costs 18\% of average monthly income across low-
and middle-income countries, rising to 51\% for the poorest quintile
\citep{gsma2024mobile}. ``Free app'' is a description of a price, not of
access.

\subsection{The Contradiction's Clearest Case: One Laptop per Child}
\label{sec:olpc}

The era's largest attempt to route around connectivity and cost was to
give out the hardware itself. The One Laptop per Child program distributed
millions of purpose-built machines at roughly \$200 per unit; Peru alone
deployed them at national scale. The program received what educational
technology rarely gets, a large randomized evaluation: across 318 rural
Peruvian schools, the intervention raised computers per student from 0.12
to 1.18, and after fifteen months produced no measurable effect on
mathematics or language test scores, alongside some positive but weaker
evidence on general cognitive skills \citep{cristia2017olpc}; a companion
randomized trial of home use likewise found little academic impact
\citep{beuermann2015olpchome}. The machines arrived; the outcomes did
not move.

We read this result carefully, because it is the hinge of the paper.
It does not show that cheap hardware is useless; it shows that hardware
without capable software, software able to do a teacher's work rather
than merely display content, does not by itself teach. In 2012 no such
software could exist: nothing that fit in those machines could read an
arbitrary text aloud, converse about it, hear the learner, or correct a
sentence. The constraint was not the device. It was what the device could
run.

\section{What Binds Low-Cost, Personalized, Equitable Language Education}
\label{sec:constraints}

Sections \ref{sec:need} and \ref{sec:tech} reduce to four constraints.
Each row of Table~\ref{tab:problems} is an instance of one of them.

\textbf{C1: the human oral provision is intrinsically expensive.} It
requires a proficient speaker's exclusive time, and both the proficiency
and the time are scarce: 44 million missing teachers, majorities of tested
teachers below the level they must teach, and class arithmetic that
dissolves individual attention (rows 1--3). Every price in
Section~\ref{sec:pay} is this constraint expressed in currency.

\textbf{C2: technological substitutes have presupposed connectivity and
recurring payment.} Broadcast, apps, and cloud AI all meter access through
a subscription, a data plan, or both, re-imposing the income gradient they
were meant to bypass (rows 6--7), and excluding the offline third of
humanity outright.

\textbf{C3: substitute content has been fixed.} Hand-engineered lessons
and app curricula cannot ingest the learner's own documents, vocabulary,
or interests; the learner adapts to the material rather than the material
to the learner (rows 4--5, 7). This is not a cosmetic limit: the need
identified in Section~\ref{sec:need} is a voice attached to \emph{the text the
learner must read}, and a fixed library, however polished, never contains
it.

\textbf{C4: capable software has never run on cheap hardware.} The one
strategy that attacked cost and connectivity directly, distributing
devices, was defeated by this constraint alone: the machines could not
run anything that taught (Section~\ref{sec:olpc}, row 8).

The four constraints are not equally durable. C1 is a fact about humans
and will not change. C2 and C3 are design choices, avoidable in principle
but economically stable as long as capable models lived only in
datacenters. C4 was, until recently, a fact about software. The next
section presents the evidence that it no longer is, and a scheme built on
its dissolution: if a machine a family already owns can run software that
performs the tutor's four roles over the learner's own documents, then C2
and C3 become unnecessary and C1 is bypassed rather than solved.

\section{The \scheme{} Scheme: A Local AI Agent for Home Language Acquisition}
\label{sec:scheme}

For readers outside artificial intelligence, we first sketch, in three short
stages, how the field arrived at models a household machine can run; readers
who know this history can proceed to Section~\ref{sec:principles}.

\subsection{From Symbolic Rules to Statistical Learning}

Turing posed the founding question, whether machines can exhibit intelligent
behavior, in 1950 \citep{turing1950computing}. For the first decades the
practical answer was hand-written rules: experts encoded knowledge as
symbolic programs, and the parser-based language tutors of Section~\ref{sec:tech}
are exactly this era's classroom form, inheriting its economics, in which
every capability is authored by specialists. Machine learning replaced
authorship with training. Gradient-based learning of internal
representations \citep{rumelhart1986learning}, convolutional networks that
read handwritten documents \citep{lecun1998gradient}, support-vector machines
\citep{cortes1995support}, and random forests \citep{breiman2001random} turned
labeled data into behavior; but each trained system still solved one narrow,
fixed task.

\subsection{Deep Learning Collapses Expert Provisions}

Deep networks scaled that recipe until breadth appeared. The 2012 ImageNet
result \citep{krizhevsky2012} showed depth plus data beating engineered
features; residual connections made very deep networks trainable
\citep{he2016deep}; recurrent memory handled sequences
\citep{hochreiter1997long}; adversarial training produced generative models
\citep{goodfellow2014generative}. What followed was a series of cost
collapses in provisions once bound to scarce experts: champion-level Go
\citep{silver2016}, dermatologist-level skin-cancer screening
\citep{esteva2017}, protein-structure prediction \citep{jumper2021},
assistance in mathematical discovery \citep{davies2021,trinh2024},
medium-range weather forecasting \citep{lam2023}, materials design
\citep{merchant2023}, robust speech recognition \citep{radford2023}, and
applied systems from urban water-infrastructure modeling
\citep{guo2024hydronet} to small-object detection in aerial imagery
\citep{tang2025retinanet}. Each collapse turned an expert service into
software; none of these systems yet conversed.

\subsection{Large Language Models and Agents}

The transformer architecture \citep{vaswani2017} made language the next
provision. Large-scale pretraining \citep{devlin2019bert}, few-shot scaling
\citep{brown2020}, and instruction tuning with human feedback
\citep{ouyang2022} produced systems that converse, explain, and correct.
The frontier of this line is closed and metered: GPT-4
\citep{openai2023gpt4}, Gemini \citep{geminiteam2023gemini}, and Claude
\citep{anthropic2024claude} are reached by subscription over a network,
which is constraint C2 restated. What matters for this paper is the
parallel open line: LLaMA \citep{touvron2023llama} began an open-weight
succession continued by Mistral \citep{jiang2023mistral}, Qwen
\citep{qwen2024qwen25}, Gemma \citep{gemmateam2024gemma2}, Phi
\citep{abdin2024phi3}, the edge-sized Llama 3.2 family
\citep{meta2024llama32}, and the reasoning-focused DeepSeek-R1
\citep{deepseekai2025r1}, models whose weights anyone may download and run
on their own machine. Coupled with reasoning-and-acting and tool-use
patterns \citep{yao2023react,schick2023toolformer}, such models become
agents: software that plans and carries out multi-step work. Applications
have moved into expert services accordingly, including clinical question
answering \citep{singhal2023}, education \citep{kasneci2023}, and
spatio-temporal signal prediction \citep{qin2025llm,yan2026llm}; our own
prior work follows the same arc, with the SouLLMate systems for
accessible, stigma-free mental-health support
\citep{guo2024soullmate,guo2024soullmateapp} and LLM-based mental-health
pre-screening \citep{tang2024advancing},  layered multi-model reasoning for long-context mental-health
assessment \citep{tang2025layered}, a collaborative MoE-LLM agent
architecture for urban pipeline monitoring \citep{guo2026aquasentinel}, conversational AI tutoring with
real-time knowledge tracing (RPKT) \citep{tang2025rpkt}, and serving LLMs
on low-resource hardware, the deployment regime this paper targets
\citep{sun2025transql+}. The scheme below spends this open, small, local
tier of the technology on the constraint analysis of
Section~\ref{sec:constraints}.

\subsection{Four Principles}
\label{sec:principles}

\textbf{P1: Web interface, fully local, lowest-cost models.} The
application is an ordinary browser interface backed by a service on the
same machine; every model is free, open-weight, and chosen to minimize
hardware demand rather than maximize benchmark scores. After a one-time
download, no network is required and nothing recurs. This deletes C2: no
subscription, no metering, no data leaving the machine, and longevity
independent of any vendor. Keeping learner data on the device also avoids a problem that is costly
to repair afterwards: removing a person's data from a model after
training, as privacy regulations such as the GDPR and the CCPA may
require, remains an open research problem whose exact solutions can
approach the cost of full retraining
\citep{guo2026unlearning,guo2026spatial}; a design under which learner
data never reach a shared model avoids the problem rather than solving
it. Running the models on the device is also what on-device machine learning
is valued for, keeping user data off the network; and because every model
in the scheme is open-weight, the model-extraction risk that a
systematization of on-device deployments documents for proprietary
models has nothing to take \citep{nayan2024sok}.

\textbf{P2: The complete four-skill loop, in the teacher's order.} The
software performs the enumerated roles of Section~\ref{sec:whatvoice}: it reads
any text aloud with the learner following (listening bound to reading);
discusses the text in spoken conversation (listening bound to speaking);
listens to the learner and gives pronunciation and grammar feedback (the
ear); and corrects the learner's written retellings (the pen). The unit of
study is the same document throughout, so the four bindings are exercised
on shared material, which is what a tutor does and what no fixed
curriculum can.

\textbf{P3: The learner's own documents and vocabulary are first-class.}
Input is whatever the learner must actually read; the vocabulary store is
built from the learner's own lookups and is deliberately re-encountered
across all four skills, in curated reading, in conversation, and in
writing feedback. This deletes C3: the material adapts to the learner
because the material \emph{is} the learner's.

\textbf{P4: AI-written, AI-understood, AI-updated.} The engineering cost
that made intelligent CALL expensive, per-language parsers and
professional maintenance (Section~\ref{sec:tech}), is itself collapsing: the
same class of models that powers the tutor can write and modify the
software that hosts it. The scheme therefore treats the codebase as an
AI-maintainable artifact: small legible modules, plug-in registries with
declarative capabilities, and behavior configured in editable prompt files
rather than compiled logic. Customization, adding a voice, a language
pair, a document format, or a pedagogical behavior, becomes a conversation
with a model rather than a contract with a vendor. Anyone able to describe
a need can have the system meet it; the reference instance of
Section~\ref{sec:instance} was itself built and is maintained this way.

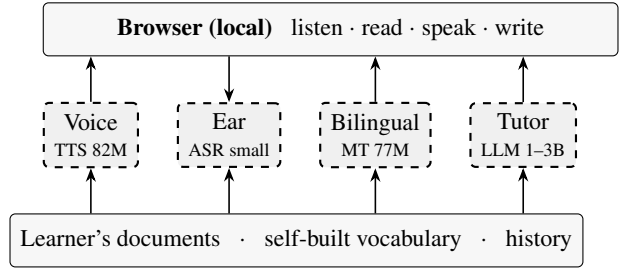
\begin{figure*}[t]
\centering
\begin{tikzpicture}[
  font=\footnotesize,
  node distance=4mm and 5mm,
  box/.style={draw, rounded corners=2pt, align=center, inner sep=4pt,
              minimum height=7mm, fill=black!3},
  plug/.style={box, dashed, thick, fill=black!6},
  arr/.style={-{Stealth[length=1.8mm]}, semithick},
]
\node[box, minimum width=76mm] (ui)
  {\textbf{Browser (local)} \; listen $\cdot$ read $\cdot$ speak $\cdot$ write};
\node[plug, below=6mm of ui.south west, anchor=north west] (tts)
  {Voice\\\scriptsize TTS 82M};
\node[plug, right=of tts] (asr) {Ear\\\scriptsize ASR small};
\node[plug, right=of asr] (mt) {Bilingual\\\scriptsize MT 77M};
\node[plug, right=of mt] (llm) {Tutor\\\scriptsize LLM 1--3B};
\node[box, below=6mm of asr.south east, anchor=north, xshift=2mm,
      minimum width=60mm] (store)
  {Learner's documents \; $\cdot$ \; self-built vocabulary \; $\cdot$ \; history};
\draw[arr] (tts.north) -- (tts.north |- ui.south);
\draw[arr] (asr.north |- ui.south) -- (asr.north);
\draw[arr] (mt.north) -- (mt.north |- ui.south);
\draw[arr] (llm.north) -- (llm.north |- ui.south);
\draw[arr] (store.north -| tts.south) -- (tts.south);
\draw[arr] (store.north -| asr.south) -- (asr.south);
\draw[arr] (store.north -| mt.south) -- (mt.south);
\draw[arr] (store.north -| llm.south) -- (llm.south);
\end{tikzpicture}
\caption{The scheme. Dashed nodes are replaceable open-weight models chosen
for minimum hardware demand; everything runs on the learner's machine over
the learner's own material. An instance is described in
Section~\ref{sec:instance}.}
\label{fig:scheme}
\end{figure*}

\subsection{Feasibility on the Cheapest Hardware}
\label{sec:feasible}

The scheme stands or falls on a factual question: can the machines that
low-cost households and schools actually have run these models? The
evidence, summarized in Table~\ref{tab:stack}, is that they can, with
room to spare.

\textbf{The stack fits.} A 1B-parameter instruction-following model
occupies 808\,MB at 4-bit quantization and a 3B model 2.02\,GB
\citep{hf2024llama32gguf}; the 82M-parameter open TTS voice weighs 327\,MB
\citep{hexgrad2025kokoro}; a Whisper-class recognizer at the ``small'' size
is under half a gigabyte on disk, with community documentation reporting
under 1\,GB resident \citep{openwhispr2026whisper,radford2023}; and compact
open translation models of the 77M class were built precisely for
real-time use on ordinary desktops and small devices
\citep{tiedemann2020,tiedemann2023opusmt,nllb2022}. The complete four-skill
stack totals under 4\,GB, inside the 8--16\,GB of the cheapest new
laptops.

\textbf{The models are capable enough.} Current 1--4B open models post
solid instruction-following and knowledge scores in their technical
reports, with the 3B class reporting MMLU in the low-to-mid 60s and strong
instruction-following benchmarks
\citep{meta2024llama32,gemmateam2024gemma2,qwen2024qwen25,abdin2024phi3};
these are not frontier systems, and the scheme does not need them to be:
the tutor's roles here are reading, discussing a provided text, and
correcting learner sentences, tasks squarely within small-model range.

\textbf{The models run at reading pace.} No peer-reviewed benchmark yet
exists for LLM inference on the cheapest CPUs, and we state throughput
only as attributed community measurement: reproducible community tables
report a 3B model at roughly 4--9 tokens per second on an \$80
single-board computer and 5--15 tokens per second on the entry laptop
CPU class \citep{localaimaster2026pi5,shah2026n100ollama,ggerganov2023m}.
The yardstick is human reading itself: silent reading averages 238 words
per minute, about four words per second \citep{brysbaert2019reading}, so
even the measured floor delivers tutor text at the pace a learner
consumes it; synthesis, recognition, and translation at the sizes above
are lighter still.

\textbf{The hardware floor is low and already owned.} New entry laptops
with 16\,GB of memory retail near \$227 in current listings (a dated
snapshot, not a market average) \citep{walmart2026n100}; the cheapest
current Apple laptop launched at \$999, \$899 for education, with 16\,GB
standard \citep{apple2025airm4}; and an 8\,GB single-board computer lists
at \$80 \citep{raspberrypi2023pi5}. For households below even this floor,
the relevant comparison is Section~\ref{sec:tech}: a device is a one-time cost
in a market where the alternative compensations recur monthly, and used
business machines circulate far below new prices.

\textbf{The marginal cost is electricity.} At the 2024 average US
residential price of 16.5 cents per kWh \citep{eia2025residential}, a
device drawing 15\,W during study, the measured envelope of entry CPUs
under inference \citep{shah2026n100ollama}, costs about a quarter of one
US cent per study hour: one cent buys four hours. Taking a generous 60\,W
full-laptop bound, the cost is still about one cent per hour, and less
wherever electricity is cheaper. After the device, the tutor's marginal
price is the light bill.

\textbf{This answers One Laptop per Child.} The randomized null result of
Section~\ref{sec:olpc} identified the missing ingredient as capable software.
That ingredient now exists, is free, is open-weight, and fits the
machines. Fitting the machines establishes feasibility, not efficacy:
whether the software teaches is a separate, measurable question, and
Section~\ref{sec:eval} equips the instance to ask it. The scheme is the
proposal to put the software there and then measure.

\begin{table}[t]
\caption{The complete four-skill stack against commodity memory. Sizes are
published file or model-card figures
\citep{hf2024llama32gguf,hexgrad2025kokoro,openwhispr2026whisper,tiedemann2020,walmart2026n100};
the resident total leaves headroom on an 8\,GB device. Throughput on the
cheapest CPUs is community-measured (see text), not peer-reviewed.}
\label{tab:stack}
\centering
\footnotesize
\setlength{\tabcolsep}{3pt}
\begin{tabular}{@{}lrr@{}}
\toprule
Component & Params & Size (Q4 / weights) \\
\midrule
Tutor LLM & 1B / 3B & 0.81 / 2.02\,GB \\
Voice (TTS) & 82M & 0.33\,GB \\
Ear (ASR, ``small'') & 244M & $\sim$0.47\,GB \\
Bilingual (MT) & 77M & $\sim$0.30\,GB \\
\midrule
Whole stack (3B tutor) & & $<$4\,GB \\
Cheapest new laptop memory & & 8--16\,GB \\
\bottomrule
\end{tabular}
\end{table}

\section{A Reference Instance: \sys}
\label{sec:instance}

\sys{} is a released, open-source implementation of all four principles:
a browser interface backed by a local service that turns any document into
a synchronized listen-read-speak-write session, plus an optional nightly
curation agent that keeps the reader supplied with personalized material.
The two programs share only a data folder; either runs without the other;
both install with two commands on Windows, macOS, or Linux, and touch the
network only for one-time model downloads and the curator's public-source
fetches. This section describes the design; consistent with the scope of
this paper, we describe what was built and how it embodies the scheme, and
we claim no learning outcomes.

\subsection{Listening Bound to Reading}

Documents (PDF, EPUB, HTML, Markdown, text) pass through a loader registry
that repairs layout, recovers ligatures, and emits a typed paragraph
stream separating body prose from headings, tables of contents, and
formula debris. The reader then plays the teacher's first role: paragraphs
are synthesized as whole units, so the voice carries the connected-speech
phenomena, linking, reduction, boundary intonation, that reading-while-
listening and shadowing practice depend on
\citep{brown2008rwl,kadota2019}; sentence boundaries are recovered from the
synthesized waveform so a highlighter tracks the spoken sentence as the
learner reads along. A translation registry renders each paragraph into
the learner's first language beneath the original, scheduled so the
sentences nearest the reading head arrive first. Selecting any word snaps
to word boundaries, resolves inflection to the dictionary head, shows IPA
and bilingual glosses, speaks the word, and files it, dated and with its
source sentence, into a self-building vocabulary that feeds a flashcard
mode; each card can be spoken by the system or pronounced by the learner for scoring.

\begin{figure*}[t]
\centering
\includegraphics[width=\textwidth]{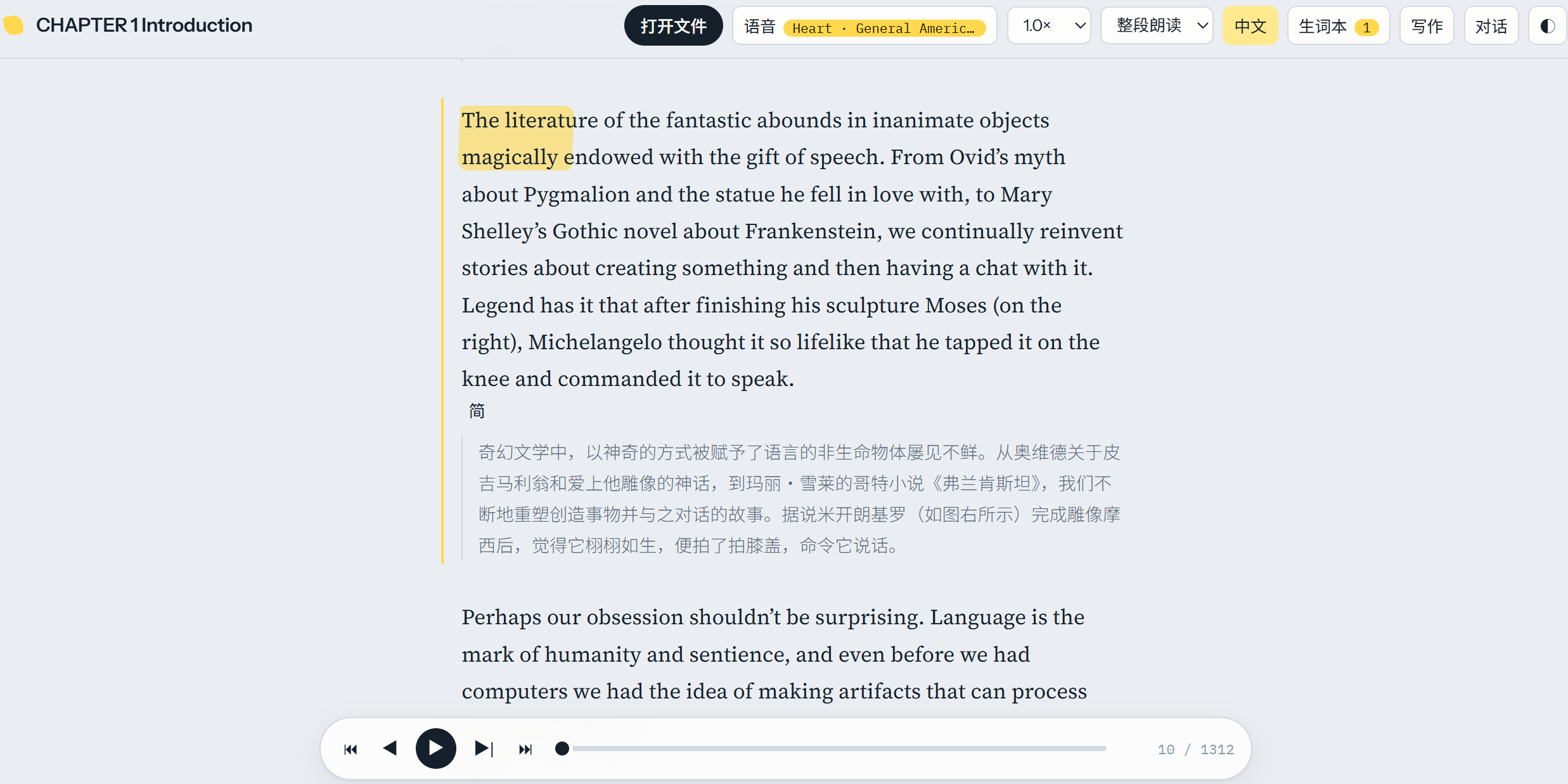}
\caption{The \sys{} reader during a read-along session, as seen by a
Chinese-speaking learner of English. Interface labels are in the
learner's first language. From left to right the toolbar reads: open
file; voice (here the Kokoro voice ``Heart'', General American); playback
speed; reading mode (whole paragraph); Chinese translation on or off;
vocabulary book, with one saved word; writing practice; conversation; and
a light or dark theme switch. The English paragraph is the text being
read aloud, and the yellow band marks the current reading position as the
voice proceeds. The grey text beneath it is a machine translation of the
same paragraph into Simplified Chinese, so the learner can check
comprehension without leaving the page. The bar at the bottom holds the
playback controls and the position in the document.}
\label{fig:reader}
\end{figure*}

Figure~\ref{fig:reader} shows a reading session in progress.

\subsection{The Ear: Segmental Pronunciation Feedback}

Following the goodness-of-pronunciation tradition \citep{wittyoung2000},
\sys{} scores a learner's recording by transcribing it with a
phoneme-level CTC model \citep{baevski2020,xu2021} and aligning the result
against a reference phoneme sequence, reporting the specific
substitutions, omissions, and insertions. The reference sequence is
obtained by transcribing the synthesized reference audio with the same
model rather than from a pronunciation dictionary, so both sequences share
one symbol inventory and one model bias and no grapheme-to-phoneme
component is required. Because the reference voice is a synthetic American
English, the aligner folds a conservative set of allophone classes (flap
and stop realizations of /t/, central-vowel variants, rhotic notation)
before counting errors, so accent-level differences are not scored as
mistakes. The feedback is deliberately segmental: it reports whether the
sounds were produced, not how native the prosody was, and the interface
says so.

\subsection{Speaking: Conversation over What Was Read}
\label{sec:talk}

The conversation module closes the loop the needs analysis identified as most rationed: someone to talk to about the text. A session opens in one of
three grounding modes, free talk, a learner-chosen topic, or the document
currently open in the reader, with document grounding following the
reading position. Interaction is push-to-talk: the learner holds a key to
speak and can interrupt the tutor at any time, a deliberate choice over
voice-activity detection for robustness on cheap microphones. Speech is
recognized locally by a Whisper-class model \citep{radford2023}, with an
interim transcription refreshed while the learner is still speaking so
words appear as they are said. After recognition, three independent jobs
run in parallel so a turn costs the slowest of them rather than their sum:
the tutor's reply (generated by the local LLM and spoken by the same
voice as the reader, with sentence-synchronized captions); a grammar-coach
pass that returns a correction card, the learner's sentence with changes
marked and a one-sentence explanation of the rule in the learner's first
language, instructed to ignore the repetitions and restarts that are
normal in speech; and the segmental pronunciation scorer above. The
tutor is prompted to deliberately reuse words from the learner's
vocabulary store; a word strip marks each stored word first when the tutor
uses it and again, fully, when the learner speaks it, making the third
re-encounter channel (heard, then said) visible.

Two design honesty notes belong in the paper rather than a footnote.
First, in free conversation the pronunciation reference is derived from
the recognizer's transcript of the learner's own utterance; this catches
distorted sounds but cannot catch a word so mispronounced that the
recognizer heard a different word, a limit the flashcard path (where the
target word is known) does not share. Second, the grammar coach operates
on the transcript, not the audio, so it corrects what was recognized, not
what was said. Both limits are stated in the interface.

\begin{figure*}[t]
\centering
\includegraphics[height=2.85cm]{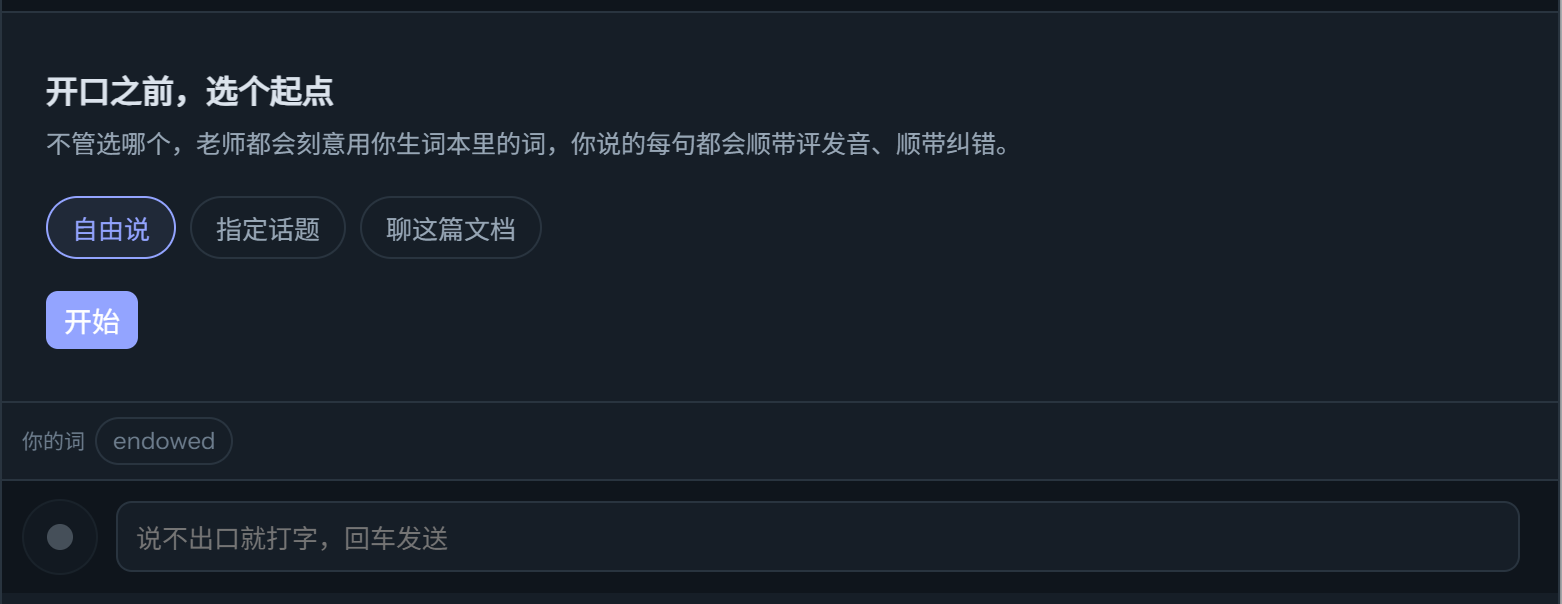}\hfill
\includegraphics[height=2.85cm]{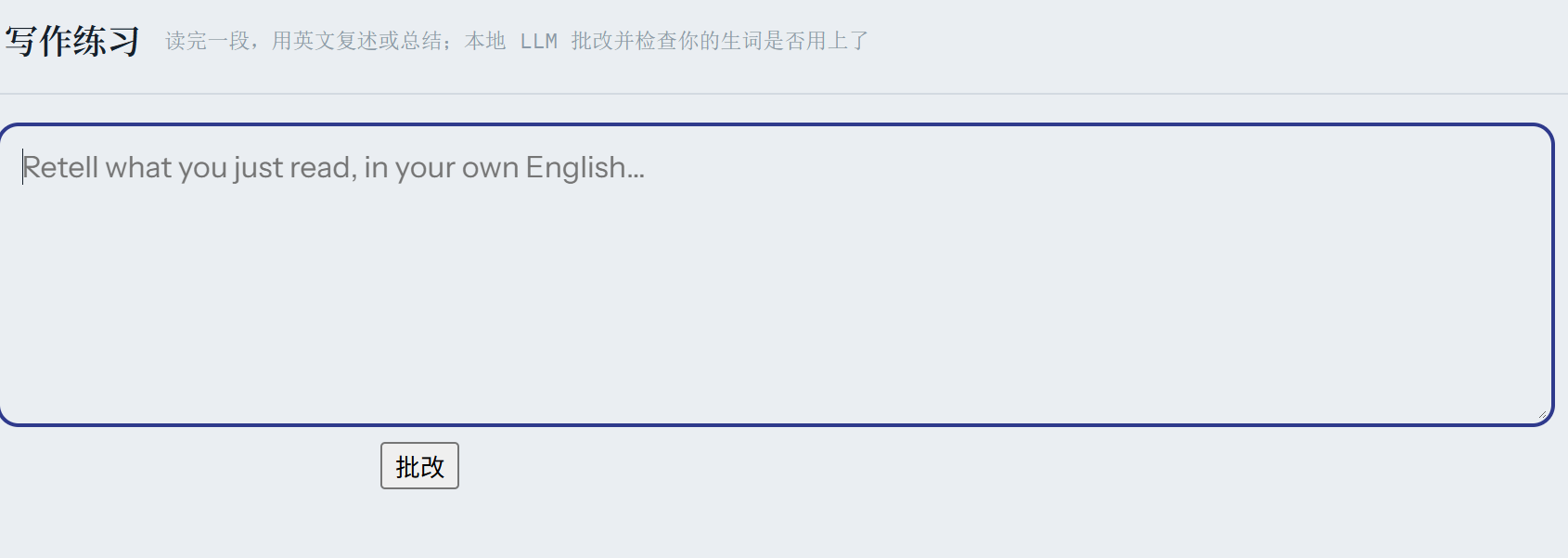}\\[2pt]
\makebox[0.48\textwidth]{\small (a) Conversation}\hfill
\makebox[0.50\textwidth]{\small (b) Writing practice}
\caption{The speaking and writing modules (interface text in Chinese,
the learner's first language). (a)~The conversation module before a
session. The heading reads ``Before you speak, choose a starting point''
and the line below it, ``Whichever you choose, the tutor will
deliberately use words from your vocabulary book, and every sentence you
say will get pronunciation feedback and corrections along the way.'' The
three options are free talk, a chosen topic, and talk about this
document, followed by a start button. The strip labelled ``your words''
holds \emph{endowed}, the word the learner saved while reading the
paragraph of Figure~\ref{fig:reader}, now carried into conversation. At
the bottom are the push-to-talk button and a text box whose placeholder
reads ``If you can't say it, type it; press Enter to send.'' (b)~The
writing panel. The header reads ``Writing practice: after reading a
section, retell or summarize it in English; the local LLM corrects it
and checks whether you used your vocabulary words.'' The button below
the text box submits the draft for correction.}
\label{fig:talkwrite}
\end{figure*}

Figure~\ref{fig:talkwrite}(a) shows the opening screen of a conversation
session.

\subsection{The Pen: Writing Practice}

After studying a section, the learner retells it in English in a writing
panel; the local model returns a corrected version with changes marked,
brief notes on the few most instructive errors, and a vocabulary check:
which of the learner's saved words appeared in the draft, and one more
that would fit, the productive counterpart of the re-encounters above. A
per-paragraph control rewrites any hard paragraph at an easier register,
technical terms kept and glossed.

Figure~\ref{fig:talkwrite}(b) shows the writing panel.

\subsection{Supply: The Night Curator}

The remaining tutor cost, choosing what to study next, is paid by an
optional second program while the machine is idle. One invocation is one
night: it fetches candidate texts from open sources on topics the learner
has rated well, ranks them partly by how densely they re-encounter the
learner's saved vocabulary, rewrites them for read-aloud delivery with
the local model (verbalizing notation, smoothing citation debris), and
writes ordinary Markdown files the reader picks up. Preference signals
are explicit ratings when given and implicit read-completion otherwise,
with explicit always overriding implicit. The curator is strictly
decoupled: it reads the reader's store read-only, keeps its own state,
and deleting it removes a convenience, never a capability.

\subsection{A Companion Tool: LLMersion Narrator}
\label{sec:narrator}

\begin{figure*}[t]
\centering
\includegraphics[height=4.6cm]{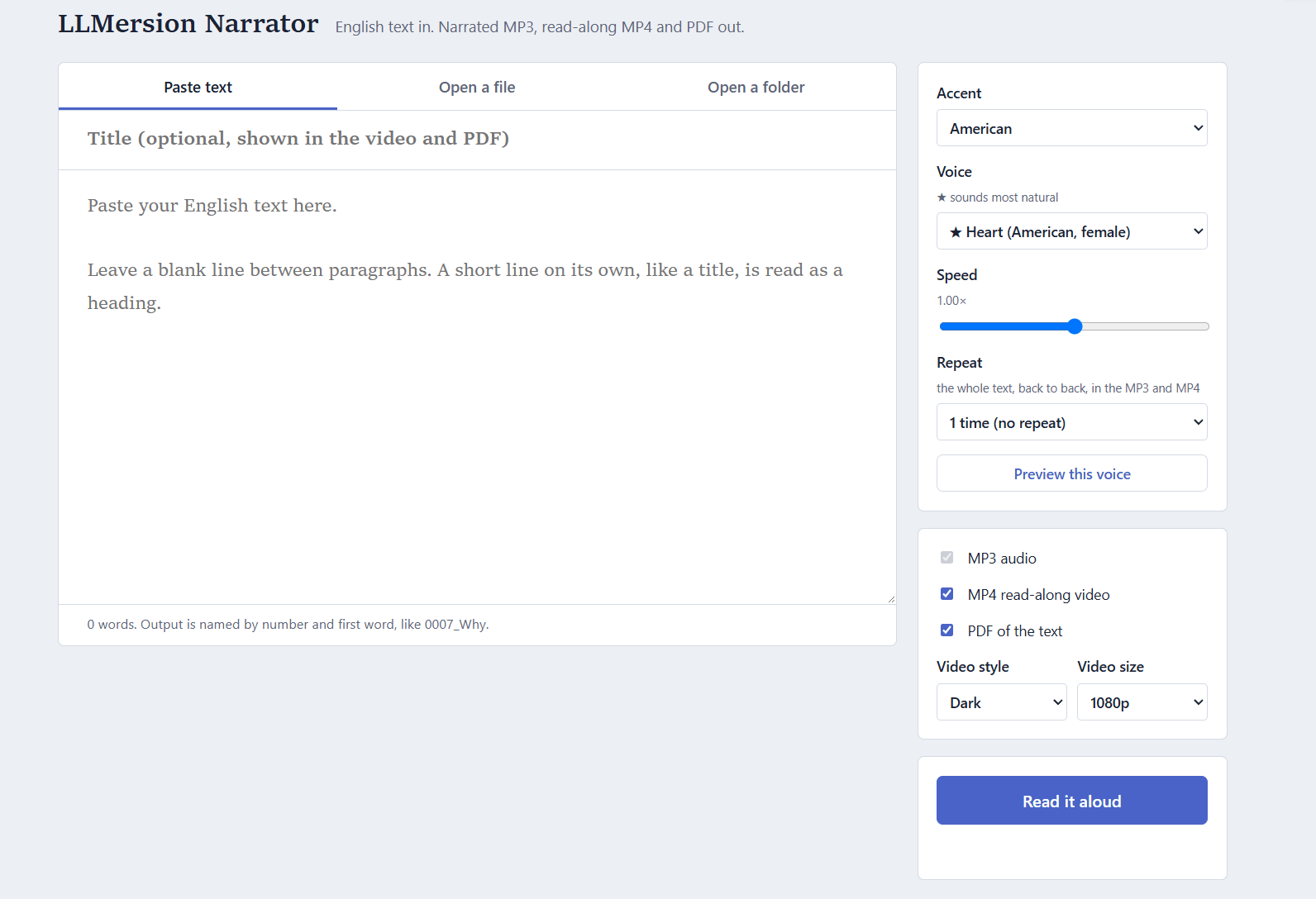}\hfill
\includegraphics[height=4.6cm]{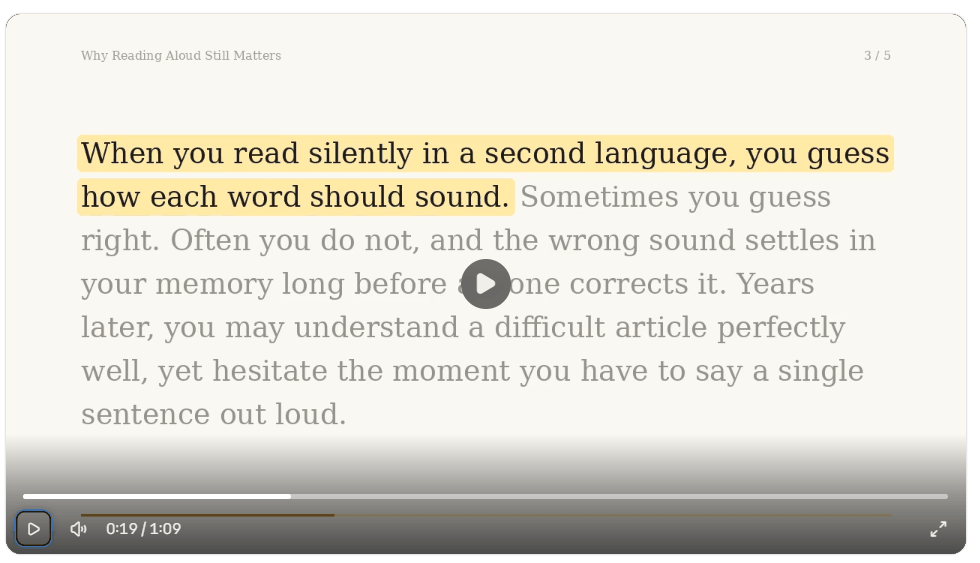}\\[2pt]
\makebox[0.45\textwidth]{\small (a) Input page}\hfill
\makebox[0.53\textwidth]{\small (b) A frame of the read-along MP4}
\caption{LLMersion Narrator, the companion tool of
Section~\ref{sec:narrator}; its interface is in English. (a)~The input
page: the learner pastes text (or uses the ``Open a file'' and ``Open a
folder'' tabs for batch work), chooses an accent, voice, speed, and
number of repetitions, selects the outputs (MP3 audio, MP4 read-along
video, PDF of the text), and presses ``Read it aloud''. (b)~A frame of
the generated video in its light style: the sentence currently being
spoken is highlighted, and the title and paragraph counter run along the
top.}
\label{fig:narrator}
\end{figure*}

The listening role is also released as a standalone companion program,
LLMersion Narrator (\url{https://github.com/QM378/llmersion-narrator}),
for learners who want results they can keep and replay away from the
computer. It reuses the reader's paragraph-level synthesis and its
waveform-derived sentence timing: the learner pastes English text, opens
a file, or opens a whole folder, and receives for each text a narrated
MP3, a read-along MP4 in which the spoken sentence is highlighted, and a
PDF of the text, optionally repeated back to back up to 20 times. Like
the reader it runs locally, and a Windows build that requires no Python
installation is distributed with the release.
Figure~\ref{fig:narrator} shows the input page and a frame of the
generated video.

\subsection{AI-Built, AI-Maintained in Practice}
\label{sec:aicode}

Principle P4 is not aspirational in the instance; it is how the system
exists. The codebase was written, debugged, and extended in AI-assisted
sessions, and its architecture is shaped for that workflow: every
substitutable component, voices, recognizers, translators, dictionaries,
document loaders, is a registry of small plug-in files sharing one
declared interface, so adding a capability is one legible file that a
model can write and a maintainer can read; pedagogical behavior, the
tutor's persona, the coach's correction policy, the curator's rewriting
style, lives in editable prompt files rather than code. The practical
consequence is the one the scheme promises: a user who wants a new voice,
another language pair, or a stricter grammar coach can describe the change
to a model, in their own language, and apply the resulting file, without
the professional engineering that priced the parser era
(Section~\ref{sec:tech}) out of individual reach. We present this as a
design principle with an existence proof, not as a quantified engineering
result; measuring its maintenance economics is future work.

\begin{figure*}[t]
\centering
\includegraphics[width=0.82\textwidth]{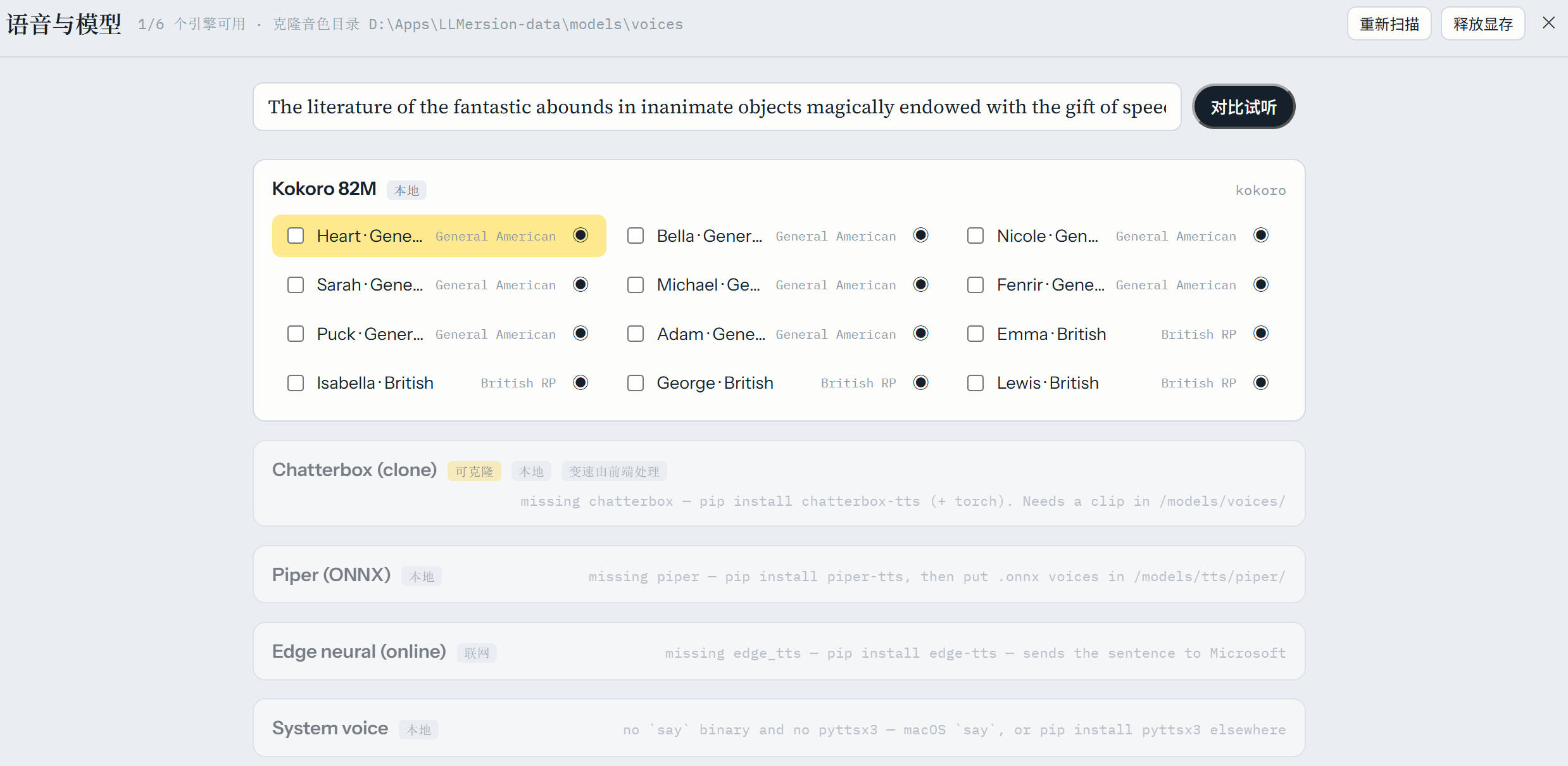}
\caption{The voice and model panel of \sys{}, an example of the plug-in
registries of Section~\ref{sec:aicode} (Chinese labels translated here).
The header reads ``Voices and models'', followed by the number of speech
engines currently usable and the folder scanned for cloned voices; the
buttons on the right rescan for engines and free GPU memory. The text
box holds a sentence from the open document, and the button beside it
plays that sentence in every checked voice for side-by-side comparison.
On this machine only the Kokoro engine (tagged ``local'') is installed,
with American and British voices. The greyed cards below are the other
registered engines: Chatterbox (tagged ``clonable'', ``local'', and
``speed handled by the front end''), Piper (``local''), Edge neural
(``online''), and the operating system's voice (``local''); each shows,
in English, the one-line installation step that would enable it.}
\label{fig:voices}
\end{figure*}

Figure~\ref{fig:voices} shows the voice registry as the learner sees it.

\subsection{Positioning and Resource Profile}

Table~\ref{tab:compare} places the instance among the tool categories
available to an individual learner; Table~\ref{tab:resources} lists the
default resident components, which sum well inside the memory of the
hardware floor of Section~\ref{sec:feasible}, with an optional larger-LLM tier
for workstation users.

\begin{table*}[t]
\caption{Feature comparison with the tool categories available to an
individual learner. Entries are design properties drawn from public
product documentation, not measured outcomes; interactive spoken feedback
from a human tutor remains the quality benchmark.}
\label{tab:compare}
\centering
\footnotesize
\setlength{\tabcolsep}{6pt}
\begin{tabular}{@{}lcccc@{}}
\toprule
 & Commercial & Cloud LLM & PDF & \sys{} \\
 & lang.\ app & chat & reader & (+curator) \\
\midrule
Learner's own documents & rarely & paste-in & \checkmark & \checkmark \\
Synchronized read-while-listen & fixed library & --- & --- & \checkmark \\
Inline translation & \checkmark & \checkmark & --- & \checkmark \\
In-context vocabulary capture & \checkmark & --- & limited & \checkmark \\
Spoken conversation on own documents & scripted & cloud tiers & --- & local \\
Pronunciation feedback & scripted drills & text advice & --- & segmental \\
Works offline & --- & --- & \checkmark & \checkmark \\
Open source, models replaceable & --- & --- & varies & \checkmark \\
Recurring service cost & subscription & metered & none & none \\
\bottomrule
\end{tabular}
\end{table*}

\begin{table}[t]
\caption{Default resident components of the reference instance
\citep{hexgrad2025kokoro,tiedemann2020,radford2023,xu2021,hf2024llama32gguf}.}
\label{tab:resources}
\centering
\footnotesize
\begin{tabular}{@{}lrr@{}}
\toprule
Component & Params & Memory \\
\midrule
TTS (Kokoro) & 82M & $\sim$0.4\,GB \\
MT (opus-mt-en-zh) & 77M & $\sim$0.3\,GB \\
ASR (Whisper base/small) & 74--244M & 0.3--0.9\,GB \\
Phoneme CTC & 0.3B & $\sim$1.2\,GB \\
Tutor LLM (quantized) & 1--4B & 0.8--2.5\,GB \\
Parsing, UI, cache & --- & $<$0.2\,GB \\
\midrule
Optional: larger local LLM & 4--27B & 3--17\,GB \\
\bottomrule
\end{tabular}
\end{table}

\subsection{Deployment Arithmetic: Memory, Pace, and Five-Year Cost}
\label{sec:arith}

The scheme-level feasibility evidence of Section~\ref{sec:feasible} can be
instantiated for \sys{}'s actual component set as three derivations. Every
input below is a cited figure; the arithmetic is ours and is stated so a
reader can recompute it.

\textbf{Memory.} Summing Table~\ref{tab:resources} for the worst case, all
components resident at once during a scored conversation turn: voice
($\sim$0.4\,GB) + translation ($\sim$0.3\,GB) + recognizer at the
``small'' size ($\sim$0.9\,GB) + phoneme scorer ($\sim$1.2\,GB) + a 3B
tutor at 4-bit quantization (2.02\,GB) + parsing, interface, and cache
($<$0.2\,GB) $\approx$ 5.0\,GB; Table~\ref{tab:stack}'s sub-4\,GB figure
is the scheme's minimal four-skill stack, and the instance's optional
phoneme scorer accounts for the difference. With a 1B tutor the same sum is
$\approx$3.8\,GB. Against the 16\,GB of the current entry-laptop
configuration \citep{walmart2026n100,apple2025airm4} this leaves a
majority of memory to the operating system and the learner's other work;
against an 8\,GB device \citep{raspberrypi2023pi5} the 1B configuration
runs comfortably and the 3B configuration fits with the phoneme scorer
loaded on demand, which is how \sys{} already treats it. No configuration
requires hardware beyond the floor devices of Section~\ref{sec:feasible}.

\textbf{Pace.} How fast must the tutor generate for education to work?
The requirement can be derived from measured human rates rather than
asserted. Adults read silently at 238 words per minute and listen to
oral delivery at 183 words per minute \citep{brysbaert2019reading}, that
is, 3.97 and 3.05 words per second; at the commonly documented ratio of
roughly 0.75 English words per model token
\citep{openai2024tokens}, these correspond to about 5.3 and 4.1 tokens
per second. The binding constraint for \sys{} is the oral one: the
tutor's reply is consumed as speech, so generation need only outrun
4.1 tokens per second. Community measurements report a 3B model at 4--9
tokens per second on the \$80 single-board floor and 5--15 on the entry
laptop class \citep{localaimaster2026pi5,shah2026n100ollama}: the floor
meets the oral requirement and the entry laptop clears the silent-reading
one. The same inputs bound the turn latency honestly: a typical
forty-word spoken reply is roughly 53 tokens, implying about 13 seconds
of generation at the floor rate before synthesis begins, halving on the
entry-laptop class; this is why the tutor is prompted toward short
conversational turns, and why the three post-recognition jobs run in
parallel (Section~\ref{sec:talk}) rather than in sequence.

\textbf{Cost.} Table~\ref{tab:fiveyear} derives the five-year cost of
one hour of daily practice througheach channel examined in the cost analysis,
native currencies preserved so no exchange-rate assumption is needed.
The scheme's recurring term is electricity alone: $1{,}825$ hours at a
generous 60\,W and the 2024 average US residential rate of 16.5 cents
per kWh \citep{eia2025residential} is about \$18; adding the dated
\$227 floor-laptop snapshot \citep{walmart2026n100} yields roughly
\$245 all-in, and a household that already owns any capable machine
pays only the \$18. Every alternative channel costs more per year than
the scheme costs per five years, and the human channels cost more per
month: the average Korean per-student spend on English tutoring alone
\citep{kosis2025private} exceeds the scheme's five-year electricity
total many times over in any recent currency terms. The comparison is
not that the alternatives are bad; Section~\ref{sec:tech} concedes they work.
It is that their price re-imposes the gradient of Section~\ref{sec:strat},
and the scheme's does not.

\begin{table}[t]
\caption{Five-year cost of one hour of daily practice, derived from
cited prices
\citep{walmart2026n100,eia2025residential,openai2023plus,italki2026pricing,kosis2025private,wei2024ciefr}.
Native currencies preserved; the device row is a dated retail snapshot,
not a market average.}
\label{tab:fiveyear}
\centering
\footnotesize
\setlength{\tabcolsep}{4pt}
\renewcommand{\arraystretch}{1.15}
\setlength{\tabcolsep}{3pt}
\begin{tabular}{@{}p{3.2cm} r p{2.3cm}@{}}
\toprule
\textbf{Channel} & \textbf{5 years} & \textbf{Basis} \\
\midrule
\sys{}, new floor laptop & $\approx$\$245 & \$227 device + \$18
electricity \\
\sys{}, owned device & $\approx$\$18 & electricity, 60\,W \\
Cloud LLM subscription & \$1{,}200 & \$20/month \\
Tutoring, 1\,h/week & \$2{,}600 & \$10/h floor \\
Korea, English tutoring & 14.9M won & 248k won/month \\
China, tutoring household & RMB 42{,}190 & RMB 8{,}438/year \\
\bottomrule
\end{tabular}
\end{table}

\subsection{A Designed Feedback Instrument}
\label{sec:eval}

The instance ships with the means of its own future evaluation, and we
state plainly that it is designed, not yet administered: no learning
outcome is claimed in this paper. A 20-item self-report instrument
(Table~\ref{tab:survey}, Appendix~\ref{app:survey}) covers reading fluency, listening, speaking and
pronunciation, vocabulary, and integrated workflow on a uniform 0--4
scale, with an optional four-item module for curator users; it is meant
for longitudinal self-comparison after the first session, one week, one
month, and one year. A conversational administration, in which a language model adapts
follow-up questions within the session, is a natural later variant
\citep{tang2025aura}; the fixed form is kept here so that responses
remain comparable across timepoints. Alongside self-report, the system logs objective
traces locally, reading throughput, pronunciation-score trajectories,
vocabulary growth, curator completion signals, which a respondent can
attach voluntarily. These traces are collected to operate the system, not designed as
comparative evidence, and data gathered for operational monitoring are
not automatically valid evaluation data \citep{lin2026compliance}; they
enter evaluation only as pre-declared endpoints of a study designed for
that purpose. Collection is designed against pre-stated
expectations rather than mined afterwards: rising subscale means across
timepoints (H1); agreement between the speaking subscale and logged
pronunciation trajectories (H2), between the vocabulary subscale and
logged list growth (H3), and between curator ratings and implicit
completion (H4); and whether the re-encounter density of a curated pick
predicts its rating and completion (H5). Establishing any of these, and
any causal claim about learning, requires the controlled study we leave
as future work; the instrument and the logged measures define its
endpoints in advance.


\section{Limitations and Outlook: Toward the Private Learning Agent}
\label{sec:outlook}

The limits of this paper are the limits of a proposal. We claim no
learning outcomes: the published evidence establishes the need and the feasibility
evidence establishes that the software runs, but whether the scheme
teaches is exactly the question the designed instrument of
Section~\ref{sec:eval} exists to ask, and it has not yet been administered.
Several feasibility inputs are honest ranges rather than settled facts:
throughput on the cheapest CPUs rests on attributed community
measurement, not peer review, and retail prices are dated snapshots.
End-to-end conversational latency, power draw, and thermal behavior on
floor hardware are likewise unmeasured; the deployment arithmetic of
Section~\ref{sec:arith} bounds the first from cited inputs but measures
none of them, and token throughput alone does not equal usable tutoring
latency. The
instance has its own stated edges: pronunciation feedback is segmental
only and, in free conversation, references the recognizer's transcript of
the learner's own speech, so a wholly misrecognized word escapes scoring;
grammar correction operates on transcripts, not audio; the resident tutor
is a small model, with a small model's occasional lapses in instruction
following; the system is single-user; and the dictionary is the last
component specific to the English-Chinese pair, though the voice,
recognition, translation, and scoring registries are already
language-agnostic.

The outlook is the scheme's name taken literally: a private learning
agent. The instance already contains the parts of one, a curator that
chooses material overnight from the learner's signals, a tutor that
discusses it by voice, a coach that corrects production, and P4 makes the
whole assembly user-modifiable; what remains is their autonomy and reach.
Concretely: difficulty modeling beyond vocabulary overlap, so curation
sequences documents rather than merely selecting them; additional
language pairs, which in the current architecture means principally a
dictionary; richer prosodic feedback for shadowing;  pooling what many learners' installations learn without pooling their
data, the setting personalized federated learning addresses for
heterogeneous and imbalanced clients \citep{xu2022fedper++}; and the controlled
study, pairing the instrument of Section~\ref{sec:eval} with the objective
traces the system already logs, that would convert this proposal's
pre-stated expectations into evidence. The direction of travel is guided by the identified constraints: every capability added must continue to run on the
machines people already have, or it re-imposes the constraints this
scheme exists to delete.

A further horizon is embodiment. A decade of research on social robots
for education finds that physical presence contributes what no screen
does, joint attention, gesture, and sustained engagement, while noting
that cost has kept such tutors out of ordinary homes
\citep{belpaeme2018social}; meanwhile embodied foundation models are
folding language, perception, and action into single systems
\citep{driess2023palme}. The scheme is built to meet that line: \sys{}'s
four-skill loop lives behind a browser and a set of model registries,
so an embodied host would replace the screen, not the stack, and the
tutor that reads, converses, listens, and corrects could inhabit
whatever low-cost embodied platforms eventually reach households. Experience with the most visible embodied deployment of AI, automated
driving, suggests that the effect of such systems depends less on the
single machine than on how it is integrated with the infrastructure
around it \citep{lin2026connected}; for a home tutor, that infrastructure is
the household's existing devices. The low-power processors such
platforms rely on are themselves an active line of open, reproducible
design work \citep{huang2026clock}. The same affordability constraint applies to embodiment: embodiment enters the
scheme when, and only when, it runs on hardware families can own.

\section{Conclusion}
\label{sec:conclusion}

The binding of listening, reading, speaking, and writing has always been
available to learners with someone to read to them, talk with them, and
correct them; the analysis in this paper documented what everyone else pays,
in tutoring bills, in widened admission gaps, in a skill profile that
reads but cannot speak. Technology's earlier compensations were real but
narrow, and the one program that put hardware directly into children's
hands showed, under randomized evaluation, that hardware alone teaches
nothing: the missing ingredient was software able to do a teacher's
work. That ingredient now fits in two gigabytes and generates at the
pace speech is consumed, on machines sold for the price of a month of
tutoring. We have proposed
a scheme for spending that fact on educational equity, four principles:
local and free, the whole four-skill loop, the learner's own material,
and a codebase its users can reshape through AI, and released \sys{} as
a working instance. The provision Bloom priced beyond societies' reach
is not beyond a household's anymore; what remains is to put it in
homes, and to measure what it does there.

\bibliography{llmersion}

\clearpage
\onecolumn
\appendix
\section{The Designed Self-Report Instrument}
\label{app:survey}

Table~\ref{tab:survey} reproduces the complete instrument of
Section~\ref{sec:eval}. It is designed and released with the system; it has
not been administered, and no response data exist at the time of
writing.

\begin{table}[h!]
\caption{The \sys{} preliminary longitudinal self-report instrument
(designed; not yet administered or validated). Mark one box per row: 0 =
strongly disagree / never, 4 = strongly agree / always. Items 21--24 apply
only to learners who use the night curator.}
\label{tab:survey}
\centering
\footnotesize
\setlength{\tabcolsep}{5pt}
\begin{tabular}{@{}r p{11.6cm} ccccc@{}}
\toprule
\multicolumn{2}{@{}l}{Time using the system:\quad
\ck~first session\quad \ck~1 week\quad \ck~1 month\quad \ck~1 year}
& 0 & 1 & 2 & 3 & 4 \\
\midrule
\multicolumn{7}{@{}l}{\itshape Reading fluency}\\
1 & I can read English technical documents for longer sessions without fatigue. & \ck & \ck & \ck & \ck & \ck \\
2 & My reading speed on technical English has increased. & \ck & \ck & \ck & \ck & \ck \\
3 & I rarely need to re-read a sentence to work out its structure. & \ck & \ck & \ck & \ck & \ck \\
4 & I rely on the translation less than when I started. & \ck & \ck & \ck & \ck & \ck \\
\multicolumn{7}{@{}l}{\itshape Listening}\\
5 & I can follow a spoken paragraph without looking at the text. & \ck & \ck & \ck & \ck & \ck \\
6 & I recognize word boundaries in fast connected speech. & \ck & \ck & \ck & \ck & \ck \\
7 & Listening at higher playback speeds is comfortable for me. & \ck & \ck & \ck & \ck & \ck \\
8 & English technical talks and videos are easier to follow than before. & \ck & \ck & \ck & \ck & \ck \\
\multicolumn{7}{@{}l}{\itshape Speaking and pronunciation}\\
9 & I read aloud (shadow) along with the voice during my sessions. & \ck & \ck & \ck & \ck & \ck \\
10 & My pronunciation scores improve on words I practice repeatedly. & \ck & \ck & \ck & \ck & \ck \\
11 & I am confident pronouncing new technical terms aloud. & \ck & \ck & \ck & \ck & \ck \\
12 & Listeners understand my spoken English without my repeating myself. & \ck & \ck & \ck & \ck & \ck \\
\multicolumn{7}{@{}l}{\itshape Vocabulary}\\
13 & I look up fewer words per page in documents from my field. & \ck & \ck & \ck & \ck & \ck \\
14 & Words I saved earlier reappear in my reading and I recognize them. & \ck & \ck & \ck & \ck & \ck \\
15 & Flashcard review helps me retain the words I saved. & \ck & \ck & \ck & \ck & \ck \\
16 & I can recall the pronunciation of saved words, not only their meaning. & \ck & \ck & \ck & \ck & \ck \\
\multicolumn{7}{@{}l}{\itshape Integrated study workflow}\\
17 & Reading with synchronized audio keeps me more focused than silent reading. & \ck & \ck & \ck & \ck & \ck \\
18 & I complete more English documents per week than before. & \ck & \ck & \ck & \ck & \ck \\
19 & I use the system for documents I genuinely need, not only practice texts. & \ck & \ck & \ck & \ck & \ck \\
20 & Overall, this system has improved my English more than my previous routine. & \ck & \ck & \ck & \ck & \ck \\
\midrule
\multicolumn{7}{@{}l}{\itshape Material supply (optional module: answer only if you use the night curator)}\\
21 & The curator's nightly picks match my interests. & \ck & \ck & \ck & \ck & \ck \\
22 & The difficulty of curated material is right for me. & \ck & \ck & \ck & \ck & \ck \\
23 & Curated material re-encounters words I have been looking up. & \ck & \ck & \ck & \ck & \ck \\
24 & I actually read the curated picks rather than letting them pile up. & \ck & \ck & \ck & \ck & \ck \\
\bottomrule
\end{tabular}
\end{table}

\end{document}